\pdfoutput=1

\documentclass[a4paper,twoside]{article}

\usepackage{epsfig}
\usepackage{subcaption}
\usepackage{calc}
\usepackage{amssymb}
\usepackage{amstext}
\usepackage{amsmath}
\usepackage{amsthm}
\usepackage{multicol}
\usepackage{pslatex}
\usepackage{apalike}
\usepackage{algorithm2e}
\usepackage[bottom]{footmisc}
\usepackage{SCITEPRESS}     
\usepackage{graphicx}
\usepackage{pdfpages}
\usepackage{booktabs}
\usepackage{array}
\usepackage[accsupp]{axessibility}

\begin{document}

\title{Vision-Language In-Context Learning Driven Few-Shot Visual Inspection Model}

\author{\authorname{
Shiryu Ueno\sup{1}\orcidAuthor{0009-0006-0842-1362}, 
Yoshikazu Hayashi\sup{1},
Shunsuke Nakatsuka\sup{1}\orcidAuthor{0000-0000-0000-0000}},
Yusei Yamada\sup{1} ,
Hiroaki Aizawa\sup{2}\orcidAuthor{0000-0002-6241-3973}, 
Kunihito Kato\sup{1},
\affiliation{\sup{1}Faculty. of Engineering, Gifu University,  1-1 Yanagido, Gifu, Japan}
\affiliation{\sup{2}Graduate School of Advanced Science and Engineering, Hiroshima University, Higashihiroshima, Hiroshima 739-8527, Japan}
\email{\{ueno, hayashi, nakatsuka, yyamada\}@cv.info.gifu-u.ac.jp, hiroaki-aizawa@hiroshima-u.ac.jp}
}


\abstract{
We propose general visual inspection model using Vision-Language Model~(VLM) with few-shot images of non-defective or defective products, along with explanatory texts that serve as inspection criteria. 
Although existing VLM exhibit high performance across various tasks, they are not trained on specific tasks such as visual inspection. 
Thus, we construct a dataset consisting of diverse images of non-defective and defective products collected from the web, along with unified formatted output text, and fine-tune VLM.
For new products, our method employs In-Context Learning, which allows the model to perform inspections with an example of non-defective or defective image and the corresponding explanatory texts with visual prompts.
This approach eliminates the need to collect a large number of training samples and re-train the model for each product. 
The experimental results show that our method achieves high performance, with MCC of 0.804 and F1-score of 0.950 on MVTec AD in a one-shot manner. 
Our code is available at~https://github.com/ia-gu/Vision-Language-In-Context-Learning-Driven-Few-Shot-Visual-Inspection-Model.
}

\onecolumn \maketitle \normalsize \setcounter{footnote}{0} \vfill

\section{Introduction}
\label{sec:intro}
\newcommand{\figone}{
\begin{figure*}[!tb]
    \begin{center}
    \includegraphics[keepaspectratio, width=\linewidth]{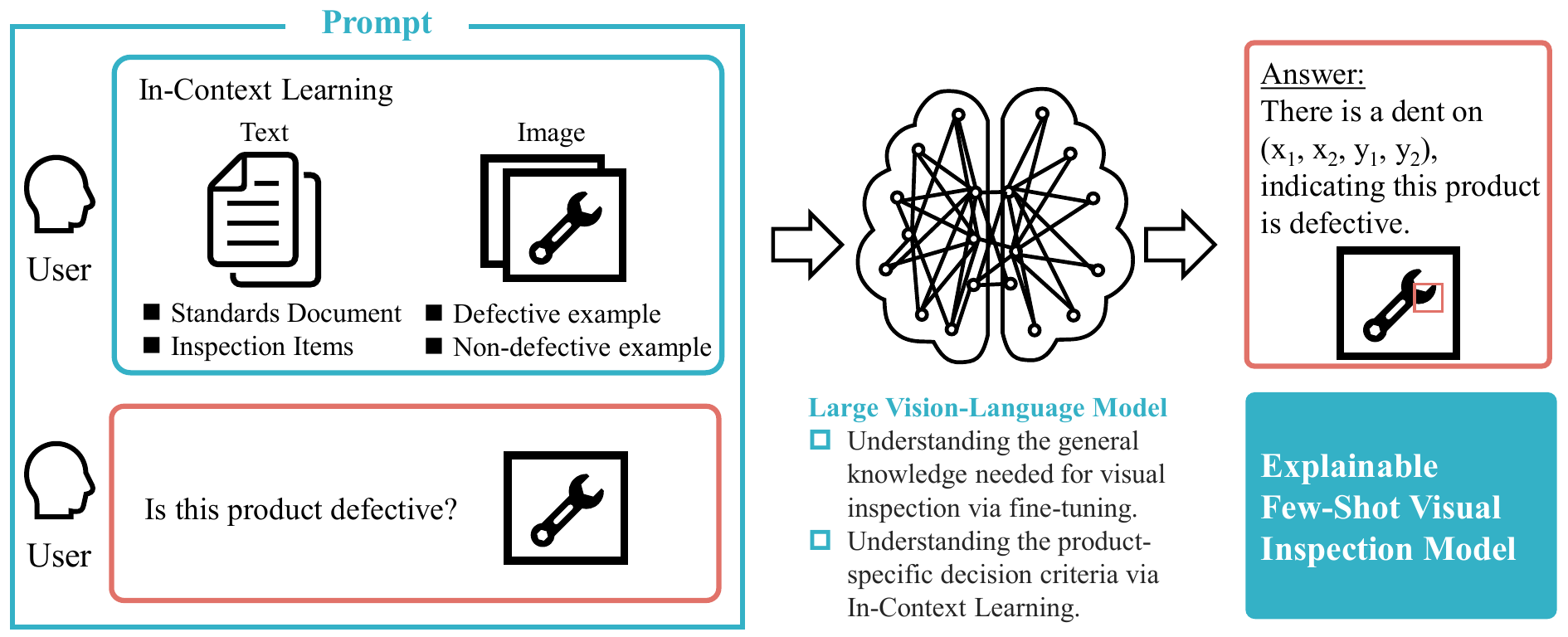}
    \caption{Framework of our proposed method. We utilize ICL for multiple image inputs to give VLM the inspection criteria of new products. Our framework gives the coordinates of the defective location, which helps the user understand the model's decision. In addition, it is easy to address by replacing the foundational model when a better VLM is proposed.}
    \label{fig1}
    \end{center}
\end{figure*}
}

\newcommand{\model}{
\begin{figure}[!tb]
    \begin{center}
    \includegraphics[keepaspectratio, width=\linewidth]{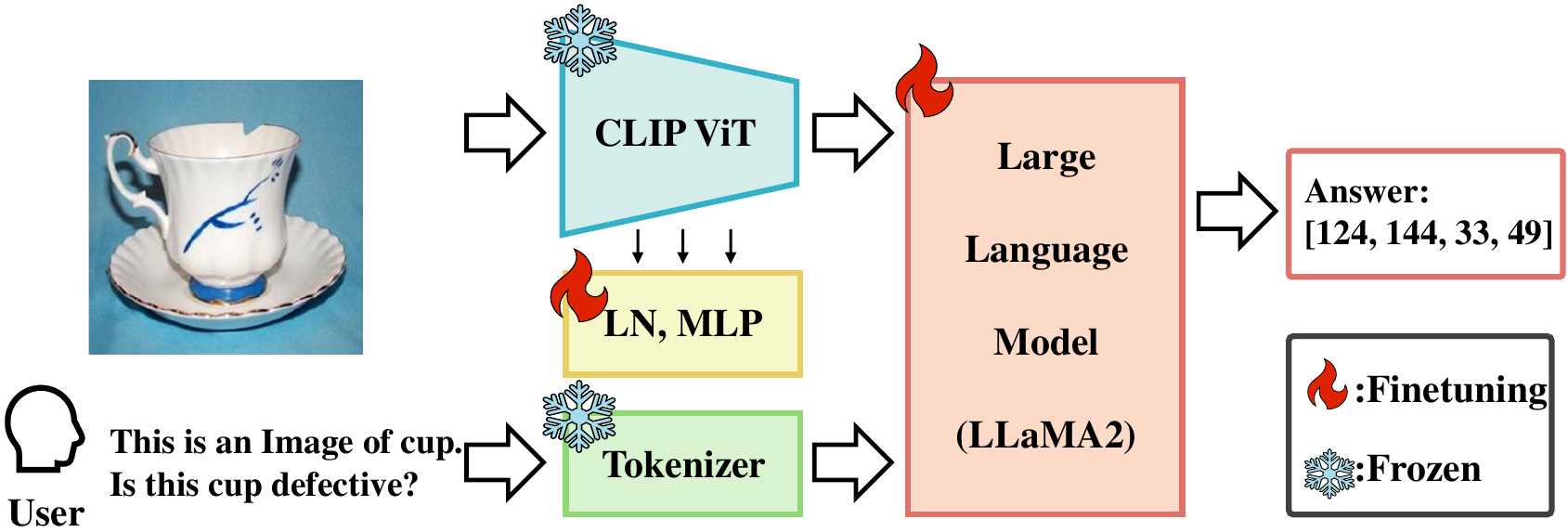}
    \caption{Architecture of ViP-LLaVA. After providing an image and the corresponding text, the image is tokenized by CLIP ViT, LayerNorm, and MLP layers, while the text is tokenized by tokenizer. Then the visual tokens and the text tokens are given to the LLM to generate the answer.}
    \label{model}
    \end{center}
\end{figure}
}

\newcommand{\collectedimages}{
\begin{figure*}[!tb]
    \begin{center}
    \includegraphics[keepaspectratio, width=\linewidth]{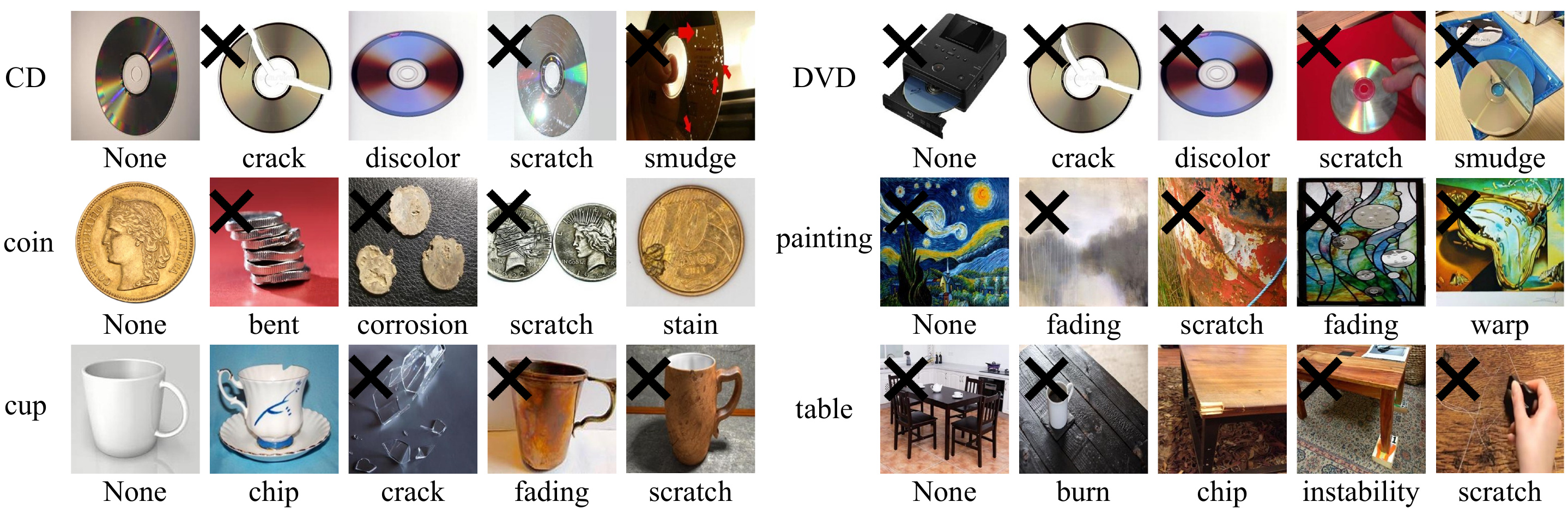}
    \caption{Examples of collected images and removal of non-defective and defective products. Images with a black cross mark are removed. 
    We remove the duplicated images and images that are too difficult for visual inspection. For instance, in ``painting'' images, there are ``fading'' or ``warp'' images, but these are considered non-defective if they are part of the image's style. Thus, we remove these kinds of images.}
    \label{collectedimages}
    \end{center}
\end{figure*}
}

\newcommand{\difficulty}{
\begin{figure}[!tb]
    \begin{center}
    \includegraphics[keepaspectratio, width=\linewidth]{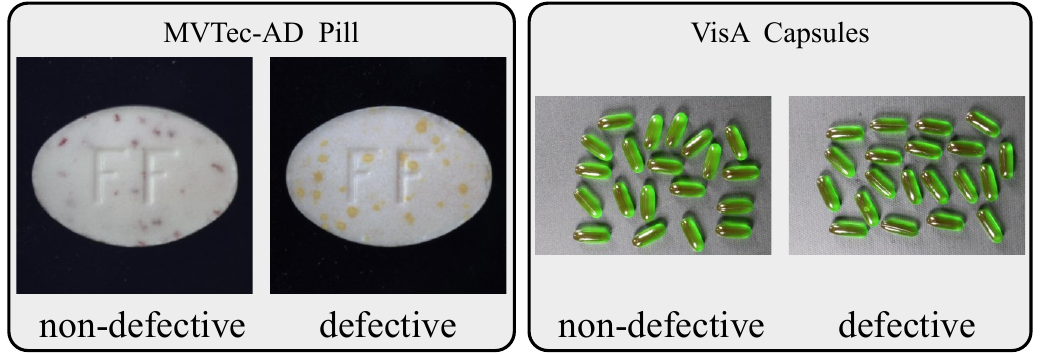}
    \caption{Examples of non-defective and defective images of ``Pill'' in MVTec AD and ``Capsules'' in VisA. For ``Pill'', the non-defective image also contains red spots, making it difficult to inspect Similarly, for ``Capsules'', the non-defective image also contains brown stains.}
    \label{difficulty}
    \end{center}
\end{figure}
}

\newcommand{\prompttrain}{
\begin{figure*}[!tb]
    \begin{center}
    \includegraphics[keepaspectratio, width=\linewidth]{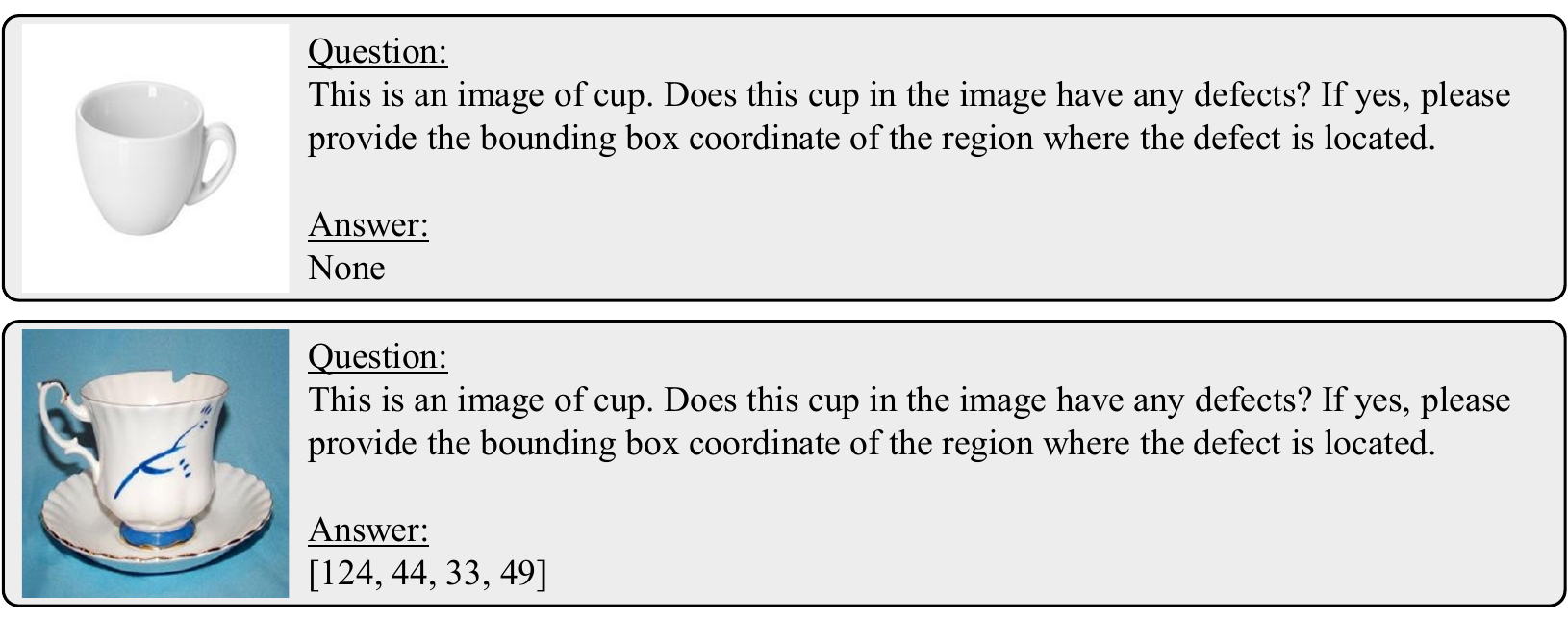}
    \caption{Prompts with unified output format. For the upper one, non-defective image, the answer is ``None''. For the lower one, defective image, the answer is the coordinates of the defective location.}
    \label{prompttrain}
    \end{center}
\end{figure*}
}

\newcommand{\prompttest}{
\begin{figure}[!tb]
    \begin{center}
    \includegraphics[keepaspectratio, width=\linewidth]{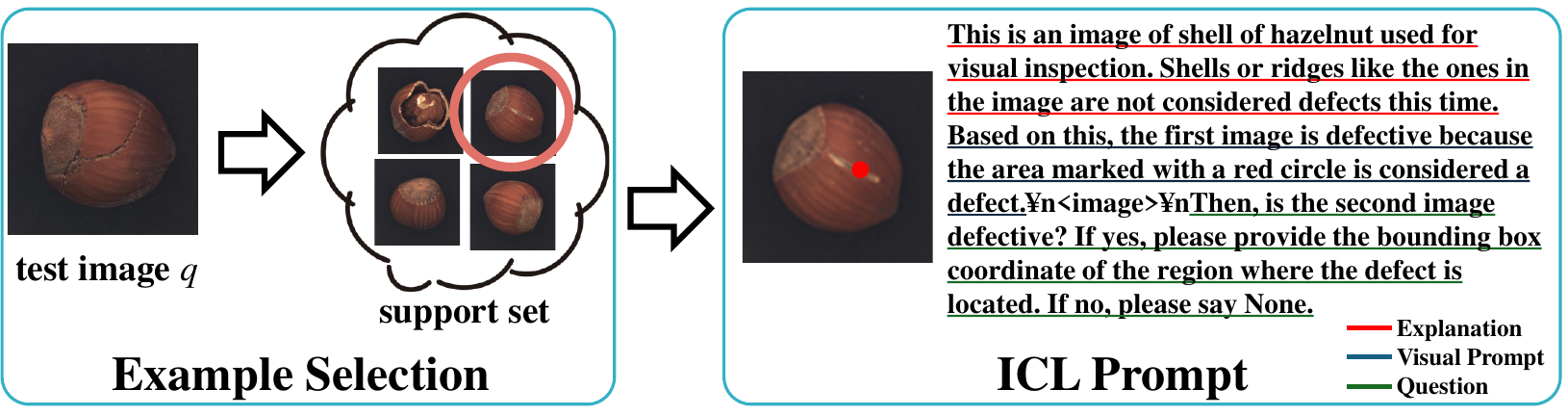}
    \caption{Framework of evaluation. First, select the example based on Eq. (1), then infer the test image with ICL.}
    \label{prompttest}
    \end{center}
\end{figure}
}

\newcommand{\mvtecbb}{
\begin{figure}[!tb]
    \begin{center}
    \includegraphics[keepaspectratio, width=\linewidth]{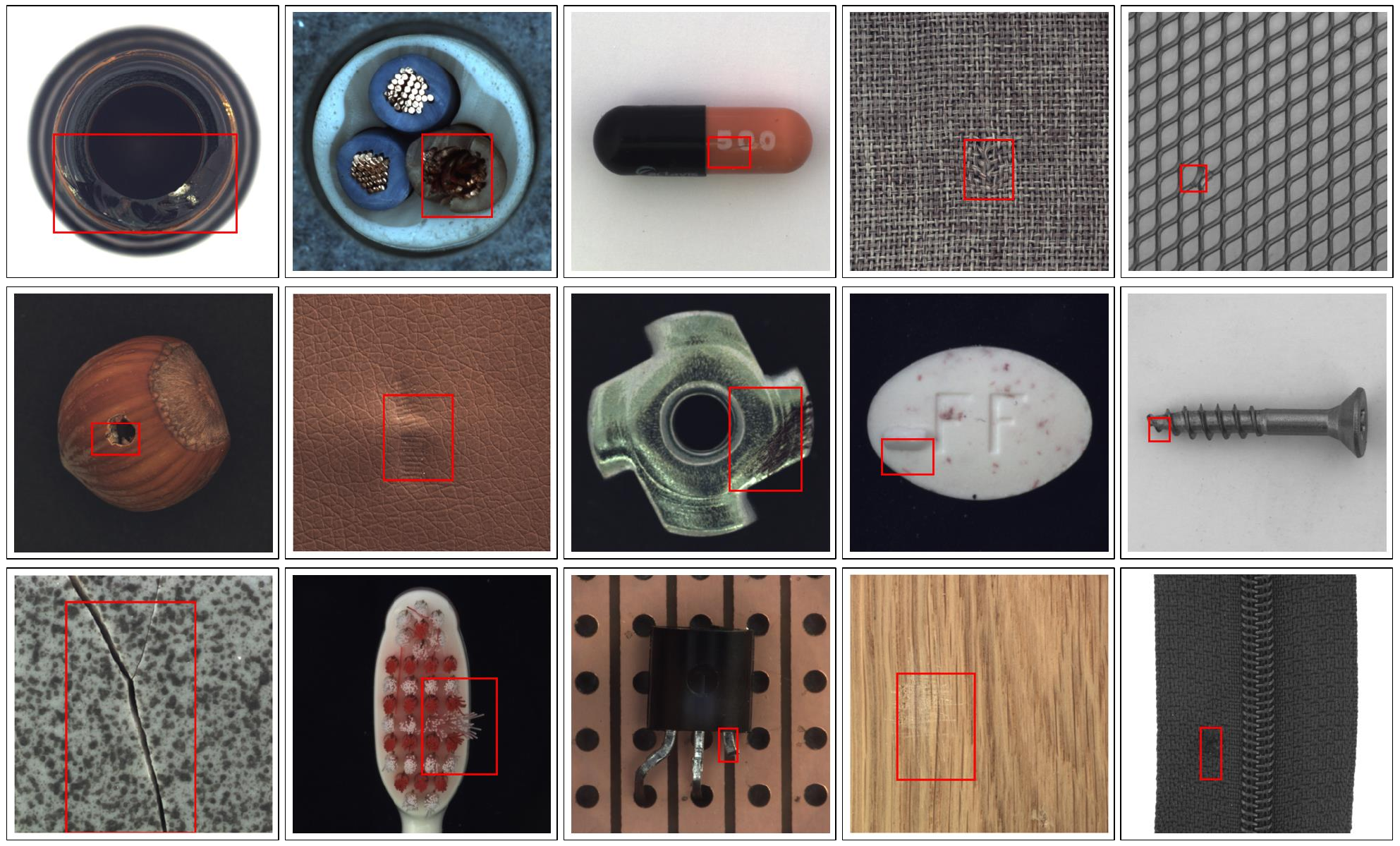}
    \caption{Visualize the model prediction for MVTec AD.}
    \label{mvtecbb2}
    \end{center}
\end{figure}
}

\newcommand{\visabb}{
\begin{figure}[!tb]
    \begin{center}
    \includegraphics[keepaspectratio, width=\linewidth]{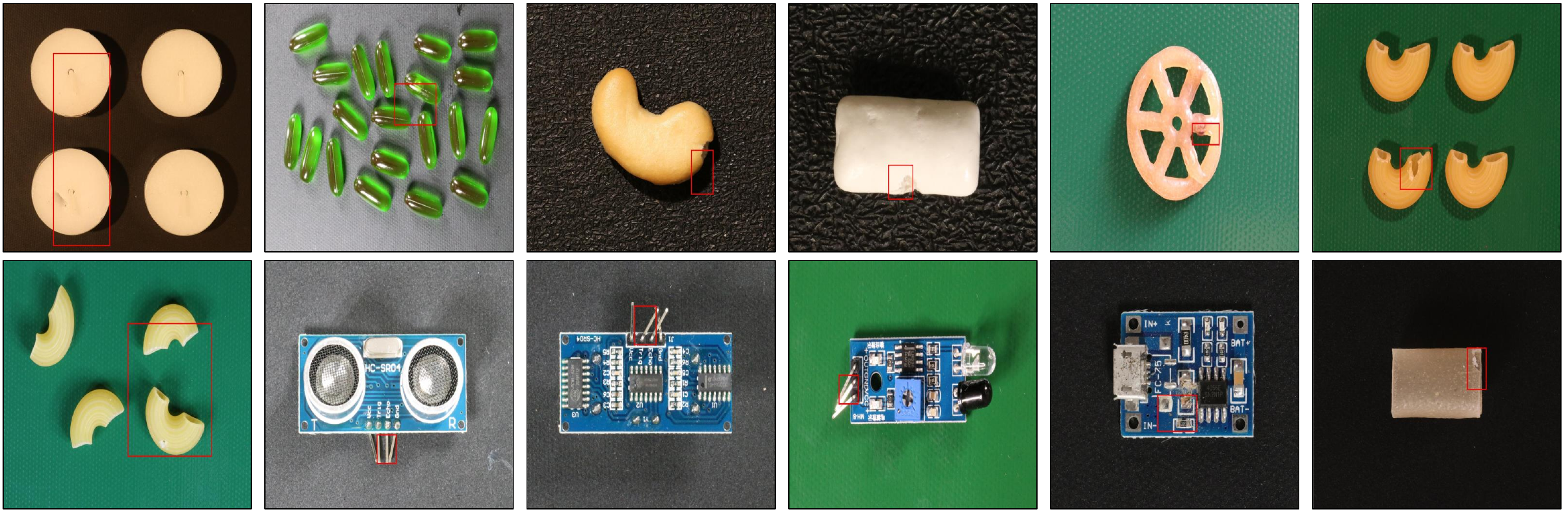}
    \caption{Visualize the model prediction for VisA.}
    \label{visa_bb}
    \end{center}
\end{figure}
}

\newcommand{\comparison}{
\begin{figure}[!tb]
    \begin{center}
    \includegraphics[keepaspectratio, width=\linewidth]{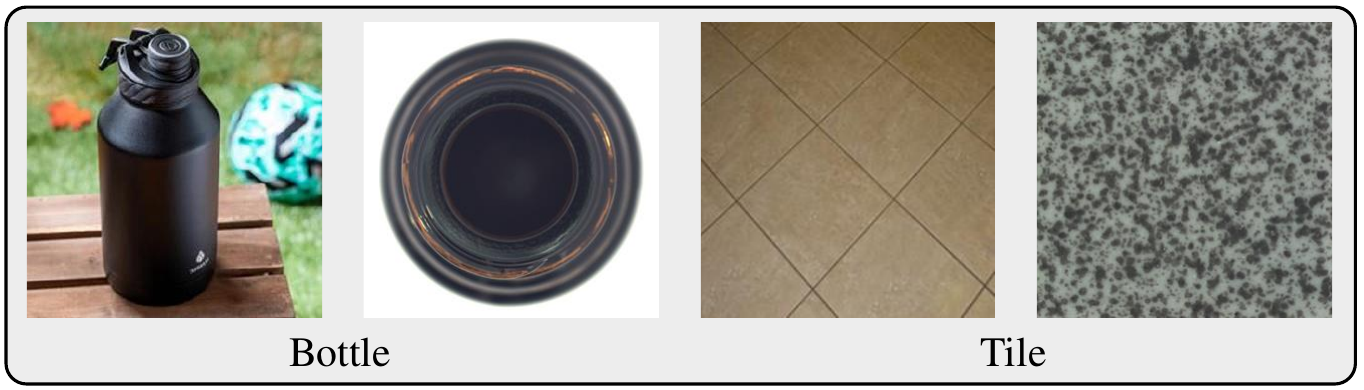}
    \caption{Examples of the images of ``Bottle'', and ``Tile'' from the collected images and MVTec AD.}
    \label{comparison}
    \end{center}
\end{figure}
}

\newcommand{\category}{
\begin{figure}[!tb]
    \begin{center}
    \includegraphics[keepaspectratio, width=\linewidth]{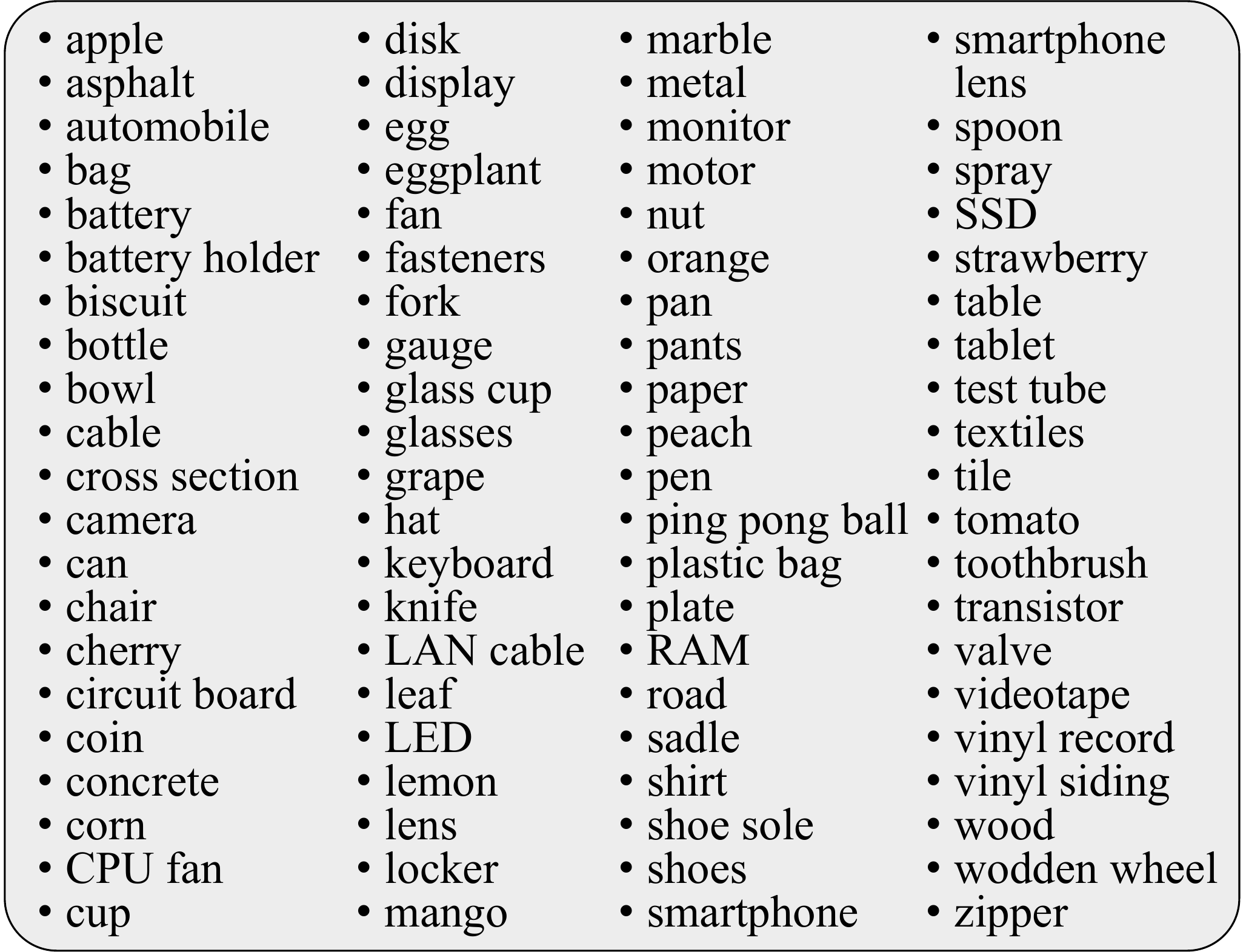}
    \caption{Product category of our dataset.}
    \label{category}
    \end{center}
\end{figure}
}

\newcommand{\resultwoft}{
\begin{figure*}[!tb]
    \begin{center}
    \includegraphics[keepaspectratio, width=\linewidth]{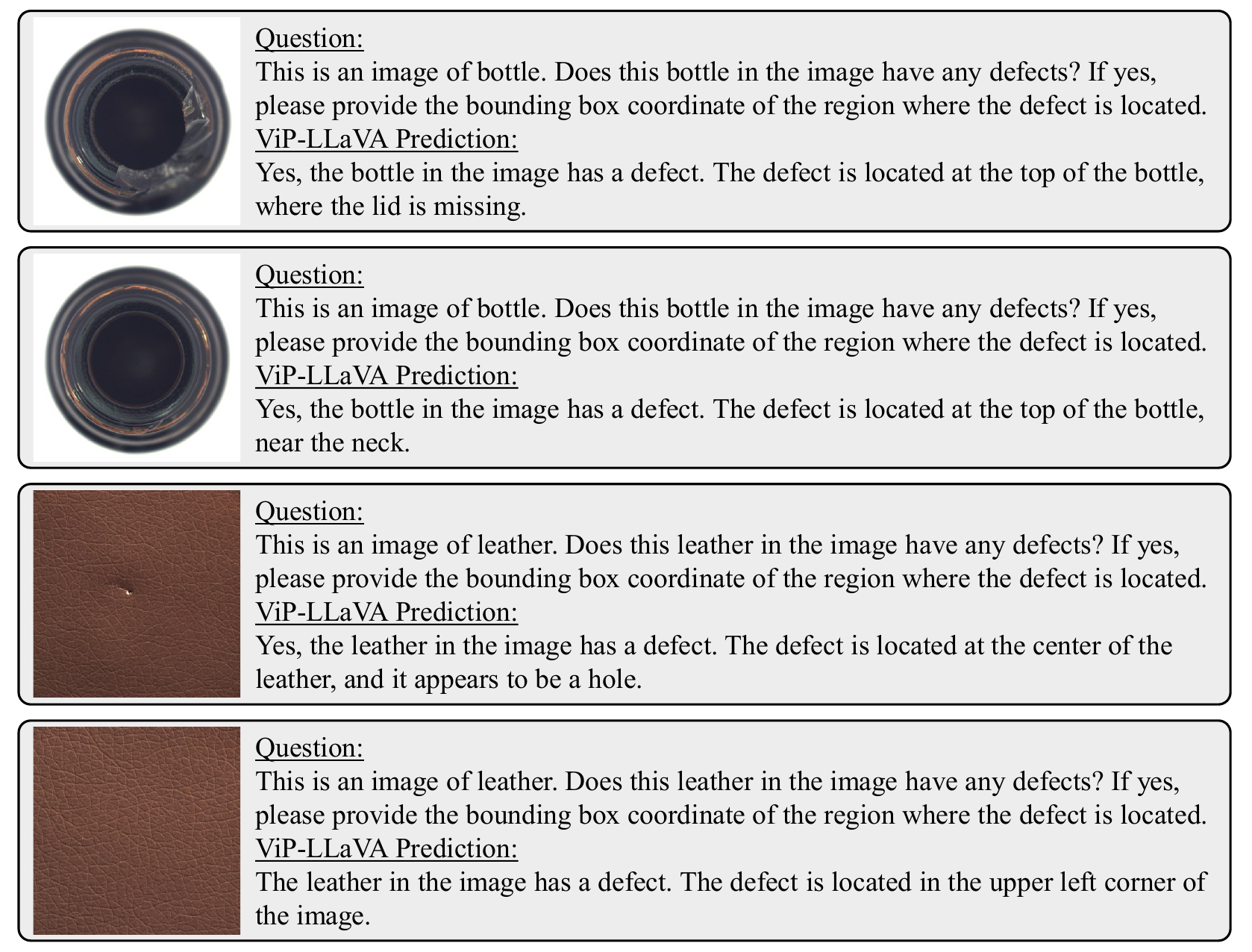}
    \caption{Result of the ViP-LLaVA before fine-tuning.}
    \label{resultwoft}
    \end{center}
\end{figure*}
}

\newcommand{\mvtecbbfull}{
\begin{figure*}[!tb]
    \begin{center}
    \includegraphics[keepaspectratio, width=0.6\linewidth]{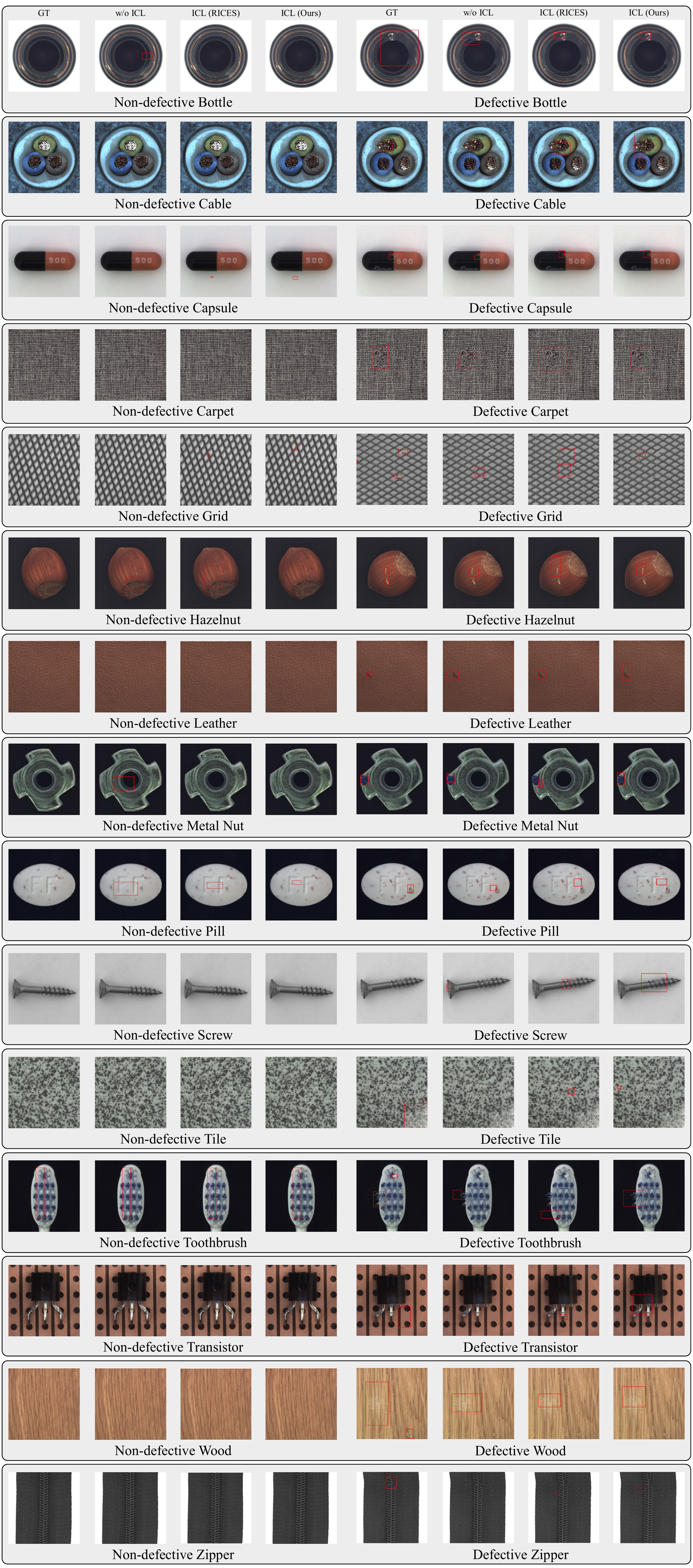}
    \caption{Visualize the model prediction of all products for MVTec AD.}
    \label{mvtecbbfull}
    \end{center}
\end{figure*}
}

\newcommand{\visabbfull}{
\begin{figure*}[!tb]
    \begin{center}
    \includegraphics[keepaspectratio, width=0.7\linewidth]{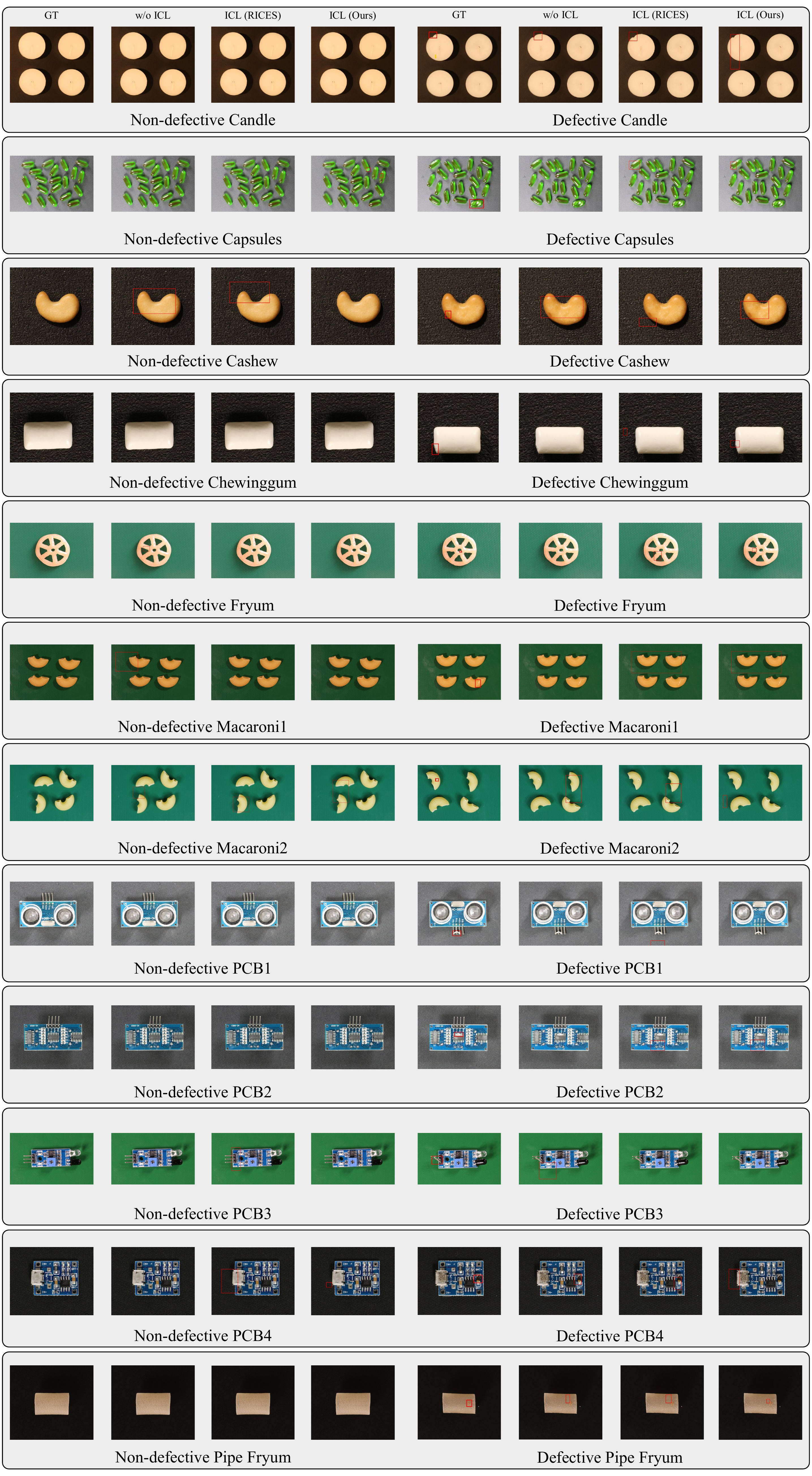}
    \caption{Visualize the model prediction of all products for VisA.}
    \label{visabbfull}
    \end{center}
\end{figure*}
}

In this study, we propose a method that can detect defective locations in new product images by using Vision-Language Model~(VLM)~\cite{SurveyMultimodalLargeLanguageModelsa}~\cite{MMBenchYourMultimodalModelAllaroundPlayer} and In-Context Learning~(ICL)~\cite{SurveyIncontextLearning}~\cite{VLICLBenchDevilDetailsBenchmarkingMultimodalInContextLearning}.

\figone

With the advancements in deep learning technology, the automation of visual inspection has become increasingly common in recent years. 
However, current visual inspection models inspect specific products by collecting a large number of images of the target product and training the model. 
Thus, these models are only applicable to the target products on which they have been trained, and re-training is necessary for new products.
Although some methods can inspect multiple products with a single model, they still require hyperparameter tuning or additional training for each product. 
In this study, we propose a general visual inspection model that leverages VLM and ICL allowing the inspection of new products without any hyperparameter tuning or model training.

Many of the current VLMs~\cite{ImprovedBaselinesVisualInstructionTuning}~\cite{ShikraUnleashingMultimodalLLMReferentialDialogueMagic} leverage Large Language Model~(LLM) to align visual and language features, demonstrating excellent performance in a wide range of tasks.
These tasks range from basic image recognition tasks, such as classification, to advanced vision-language tasks, such as Visual Question Answering~(VQA). 
However, these VLMs are not trained on specific tasks such as visual inspection.

In this study, we propose a general visual inspection model that can detect defective locations in new products without any hyperparameter tuning or model re-training, using VLM and ICL.
The framework of our proposed method is shown in Fig.~\ref{fig1}. 
First, we fine-tune the VLM for general visual inspection with a dataset constructed from a diverse set of non-defective and defective product images collected from the web.
In this study, we use ViP-LLaVA~\cite{ViPLLaVAMakingLargeMultimodalModelsUnderstandArbitraryVisualPrompts}, which has been trained on visual prompt recognition, as the foundation of our VLM, and fine-tune it with our dataset.
In addition, in typical visual inspection processes by humans, inspectors use inspection standards for the target products.
To emulate this inspection process by human, we use ICL during the evaluation to provide an example of non-defective or defective product image along with explanatory texts that serve as inspection criteria.
ICL is a method the model learns from few-shot input-output examples as prompts, without parameter update.
Using ICL during the inference of new products, we provide VLM with inspection criteria, enabling specific inspection of target products.
Since ICL performance varies significantly based on the provided examples, we propose an algorithm that can select high-quality example based on the distance in Euclidean space. 
Consequently, our proposed method does not need to collect a large number of images or to re-train the model for each target product.

In summary, our main contributions are:

\begin{itemize}
\item We propose a general visual inspection model capable of inspections and detecting defective locations for new products using VLM and ICL with only an example.
In our proposed method, fine-tune VLM on visual inspection and utilize ICL enabling the inspection of specific products.
\item We construct a new dataset consisting of diverse non-defective and defective products collected from the web, along with unified formatted output, for fine-tuning.
Also, our dataset includes coordinates of defective locations for defective products, ensuring the explainability of the model.
\item To empirically verify the proposed methodology, we evaluate on MVTec AD~\cite{MVTecADComprehensiveRealWorldDatasetUnsupervisedAnomalyDetection} and VisA~\cite{SPottheDifferenceSelfSupervisedPretrainingAnomalyDetectionSegmentation}. 
Our method achieves MCC~\cite{advantagesMatthewscorrelationcoefficientMCCF1scoreaccuracybinaryclassificationevaluation} of 0.804 and F1-score~\cite{AccuracyFScoreROCFamilyDiscriminantMeasuresPerformanceEvaluation} of 0.950 on the MVTec AD dataset in a one-shot manner.
\end{itemize}

\section{Related Work}
\label{sec:relatedwork}

\subsection{Visual Inspection}
\label{sec:vi}
Many visual inspection methods based on deep learning are trained only on non-defective images~\cite{PatchSVDDPatchlevelSVDDAnomalyDetectionSegmentation}~\cite{Padimpatchdistributionmodelingframeworkanomalydetectionlocalization}.
Thus, such methods require the collection of training samples and the re-training of the model for each target product. 
Consequently, it is challenging to apply the same model to different products without re-training.

Recently, visual inspection methods combining vision and language have been proposed.
AnomalyGPT~\cite{AnomalyGPTDetectingIndustrialAnomaliesUsingLargeVisionLanguageModelsa} can detect defective locations by learning an image decoder from non-defective and pseudo-defective images. 
However, AnomalyGPT utilizes PaDiM or PatchCore~\cite{totalrecallindustrialanomalydetection} for anomaly maps, and these methods need re-training for each products.
WinCLIP~\cite{WinCLIPZeroFewShotAnomalyClassificationSegmentation} calculates the similarity between images and texts of non-defective and defective images using CLIP~\cite{LearningTransferableVisualModelsNaturalLanguageSupervisiona} and can detect defective locations by using relative anomaly scores.
However, WinCLIP only assigns anomaly scores to test samples during inference. 
To inspect correctly, it is necessary to experimentally determine the optimal threshold on test samples.
Thus, these existing approaches cannot be considered general visual inspection models.

\subsection{Vision-Language Model}
\label{sec:VLM}
VLMs leverage LLM to align visual and language features, demonstrating excellent performance across a wide range of tasks, from basic image recognition tasks such as classification, to advanced vision-language tasks, such as VQA. 
For example, LLaVA~\cite{VisualInstructionTuning} inputs the vision embedding vectors and language embedding vectors into the LLM decoder to learn the alignment between vision and language.
LLaVA has spawned many derivative methods, among which ViP-LLaVA focuses on visual prompt recognition by utilizing a dataset where arrows or visual cues are directly embedded in the input images, thereby strengthening the alignment between low-level image details and language. 
However, these VLMs have not been trained on visual inspection tasks and thus lack the general knowledge for visual inspection~\cite{MMBenchYourMultimodalModelAllaroundPlayer}.

\model

\subsection{In-Context Learning}
\label{sec:icl}
ICL is a method that the model learns from few-shot input-output examples as prompts, without updating model parameters.
For instance, given the input ``Example input: (4, 2), Example output: 6, Question: (5, 6),'' the model infers from the provided example that the task is addition and can answer ``11.''
In multi-modal ICL, the model makes inferences based on images, prompts, and their examples.
Many VLMs are trained on diverse image-text pairs, enabling them to acquire ICL capabilities~\cite{UnderstandingImprovingInContextLearningVisionlanguageModels}.

Some VLMs are explicitly built to enhance ICL capabilities.
Otter~\cite{OtterMultiModalModelInContextInstructionTuninga} enhances ICL capabilities by fine-tuning Open Flamingo~\cite{OpenFlamingoOpenSourceFrameworkTrainingLargeAutoregressiveVisionLanguageModelsa} on MIMIC-IT~\cite{MIMICITMultiModalInContextInstructionTuninga}, which is in an ICL and Instruction Tuning format. 
At the same time, not to forget the knowledge of Open Flamingo, Otter only update parameters of Perceiver Resampler and Cross Attention Layer in language model.
Similarly, LCL~\cite{LinkContextLearningMultimodalLLMsa} proposes a new evaluation dataset, ISEKAI, which includes new concepts in the examples, making it challenging without seeing the examples.
To address ISEKAI, LCL enhances its ICL capability by fully fine-tuning Shikra~\cite{ShikraUnleashingMultimodalLLMReferentialDialogueMagic} on a custom dataset based on ImageNet~\cite{Imagenetlargescalehierarchicalimagedatabase}.
However, in practice, these VLM explicitly designed to enhance ICL capabilities do not necessarily outperform regular VLM~\cite{UnderstandingImprovingInContextLearningVisionlanguageModels}~\cite{VLICLBenchDevilDetailsBenchmarkingMultimodalInContextLearning}.

\section{Proposed Method}

\subsection{Overview}
\label{sec:overview}
In this study, we propose a general visual inspection model that combines VLM and ICL, enabling the specific inspection of new products without parameter optimization.
In addition, by constructing unified output format dataset for fine-tuning, we enable quantitative evaluation of visual inspections using VLM. 
An overview of the proposed method is shown in Fig.~\ref{fig1}.

\subsection{Model}
\label{sec:model}
In this study, we use ViP-LLaVA~\cite{ViPLLaVAMakingLargeMultimodalModelsUnderstandArbitraryVisualPrompts} as the foundational VLM.
ViP-LLaVA is a model that improves recognition capabilities for visual prompts by fine-tuning LLaVA 1.5~\cite{ImprovedBaselinesVisualInstructionTuning} on a dataset where red circles or arrows are overlaid on the original images. 
In addition to this, ViP-LLaVA utilizes the multi-level visual features to address the tendency of CLIP’s deeper features to overlook low-level details.
These methodologies improves the recognition capability for low-level details, which is especially needed for visual inspection.
ViP-LLaVA has not been trained on visual inspection tasks.

The model architecture of ViP-LLaVA is shown in Fig.~\ref{model}. 
ViP-LLaVA consists of a vision encoder to extract visual features, LayerNorm~\cite{LayerNormalization} and an MLP to tokenize visual features, a tokenizer to tokenize the language, and an LLM to generate text from these tokens.
The vision encoder is CLIP-ViT-L/14\cite{LearningTransferableVisualModelsNaturalLanguageSupervisiona}, and the LLM is LLaMA2~\cite{Llama2OpenFoundationFineTunedChatModels} 
During fine-tuning, we update the parameters of the LayerNorm, MLP, and LLM in accordance with the ViP-LLaVA procedure.

\subsection{Dataset for Fine-tune}
\label{sec:dataset}
To enhance the general knowledge of existing VLM for visual inspection, we collect images of non-defective and defective products from the web. The image collection process consists of five main steps:

\begin{enumerate}
\item Generate product names and inspection-related keywords~(e.g., ``disk'', ``broken disk'', ``discolored disk'') by GPT-4~\cite{GPT4TechnicalReport}.

\item Expand the keywords into eight languages: English, Chinese, Spanish, French, Portuguese, German, Italian, and Japanese.

\item Perform image searches using the expanded keywords and collect images with selenium.

\item Remove duplicate or unclear images.

\item Annotate the defective location coordinates for the defective images in the remaining set.
\end{enumerate}

\difficulty

Through this procedure, we collect images of various products. Each product category includes images of non-defective and defective products with up to five types of defects. 
Finally, we obtained a final set of 941 images of 84 categories.

After collecting the images, we construct a dataset for fine-tuning.
The format of the dataset is based on VQA~(i.e., a pair of question and answer for each images).
Question is ``This is an image of \{product\}. Does this \{product\} in the image have any defects? If yes, please provide the anomaly mode and the bounding box coordinate of the region where the defect is located. If no, please say None.'', answer is coordinates of the defective location for defective image, ``None'' for non-defective image.

\subsection{In-Context Learning Driven Visual Inspection}
\label{sec:iclevaluation}
It is challenging to inspect new products from a single image. 
An example is shown in Fig.~\ref{difficulty}. 
As shown in Fig.~\ref{difficulty}, some products need their specific inspection criteria for accurate visual inspection.
Thus, in this study, we utilize ICL to provide an example of non-defective or defective image along with explanatory texts that serve as inspection criteria. 
Based on the example, model precisely inspects new products.

In addition, in multi-modal ICL, example images significantly influence the output of VLM~\cite{WhatMakesMultimodalInContextLearningWork}.
RICES~\cite{EmpiricalStudyGPT3FewShotKnowledgeBasedVQA} is an existing algorithm for selecting examples in ICL, it uses cosine similarity of features.
However, cosine similarity can yield high values when the scales of features differ or when the feature dimensions are large, failing to accurately evaluate similarity~\cite{CosineSimilarityEmbeddingsReallySimilarity}.
Thus, in this study, we propose a new selection algorithm. Our proposed method is shown in Eq.~(\ref{eq:ours}).

\begin{equation}
    \operatorname{argmin} \left( \min \left\| f(x_i) - f(x_q) \right\|_2 \right)
  \label{eq:ours}
\end{equation}

Where $x$ denotes the image, $f$ denotes the vision encoder~(pre-trained ResNet50~\cite{DeepResidualLearningImageRecognition}), and $q$ denotes the index of the test image for inference, $i$ denotes the index of the image except for $q$.
Eq.~(\ref{eq:ours}) is an algorithm that selects neighboring image of the test image as example based on Euclidean distance.
By this, Eq.~(\ref{eq:ours}) takes into account the scale and dimensions of the features, and expected to select better example compared to RICES.

\section{Experiment}
\subsection{Settings}
\label{sec:settings}

The dataset used for fine-tuning was collected and created as described in Sec.~\ref{sec:dataset}. 
We perform one-stage fine-tuning for 300 epochs using ZeRO2 with XTuner~\cite{XTunerToolkitEfficientlyFinetuningLLM} on 8 NVIDIA 6000Ada-48GB GPUs. The batch size is set to 4, thus the global batch size is set to 32.
We utilized the AdamW~\cite{DecoupledWeightDecayRegularizationa} optimizer and 1e-4 learning rate, with the warm-up ratio set to 0.03. We also apply cosine decay to the learning rate.

To evaluate the performance, we used MVTec AD and VisA. These datasets were not used at all during training.
During the evaluation, one example image is given by using ICL~(i.e., one-shot manner). The method for selecting the example image follows the procedure described in Eq.~(\ref{eq:ours}). 
Also, to evaluate the effectiveness of the proposed example image selection algorithm, we compare its performance without example image and with that of RICES.
Finally, framework of the evaluation is shown in Fig.~\ref{prompttest}.

\prompttest
\begin{table*}[!tb]
\caption{Result of MVTec-AD. 'N/A' means that zero division occurred. Bold means the highest performance.}
\label{tab:result_mvtec}
\centering
\begin{tabular}{lcccccccc}
\toprule
\multicolumn{1}{c}{\textbf{Settings}} & \multicolumn{2}{c}{\textbf{Vanilla}} & \multicolumn{2}{c}{\textbf{w/o ICL}} & \multicolumn{2}{c}{\textbf{ICL (RICES)}} & \multicolumn{2}{c}{\textbf{ICL (Ours)}} \\
\cmidrule(r){1-9}
\multicolumn{1}{c}{Product Name} & F1-score & MCC & F1-score & MCC & F1-score & MCC & F1-score & MCC \\
\midrule
Bottle & N/A & N/A & 0.863 & N/A & 0.892 & 0.510 & 0.917 & \textbf{0.610} \\
Cable & N/A & N/A & 0.400 & 0.338 & 0.795 & 0.564 & 0.899 & \textbf{0.754} \\
Capsule & N/A & N/A & 0.750 & 0.426 & 0.912 & 0.384 & 0.946 & \textbf{0.658} \\
Carpet & 0.044 & 0.074 & 1.000 & \textbf{1.000} & 0.983 & 0.929 & 1.000 & \textbf{1.000}  \\
Grid & N/A & N/A & 0.973 & 0.910 & 0.884 & 0.476 & 0.982 & \textbf{0.935} \\
Hazelnut & 0.228 & 0.226 & 0.780 & N/A & 0.795 & 0.257 & 0.800 & \textbf{0.289} \\
Leather & 0.043 & 0.076 & 1.000 & \textbf{1.000} & 1.000 & \textbf{1.000} & 1.000 & \textbf{1.000} \\
Metal Nut & N/A & N/A & 0.832 & 0.540 & 0.912 & 0.468 & 0.989 & \textbf{0.947} \\
Pill & N/A & N/A & 0.838 & 0.402 & 0.922 & 0.368 & 0.968 & \textbf{0.814} \\
Screw & 0.209 & 0.140 & 0.851 & 0.244 & 0.903 & 0.506 & 0.925 & \textbf{0.673} \\
Tile & N/A & N/A & 0.957 & 0.870 & 0.977 & \textbf{0.916} & 0.977 & \textbf{0.916} \\
Toothbrush & N/A & N/A & 0.906 & 0.633 & 0.866 & 0.418 & 0.921 & \textbf{0.697} \\
Transistor & N/A & N/A & 0.762 & 0.592 & 0.871 & 0.780 & 0.894 & \textbf{0.821} \\
Wood & N/A & N/A & 0.976 & 0.752 & 0.992 & \textbf{0.965} & 0.992 & \textbf{0.965} \\
Zipper & N/A & N/A & 0.741 & 0.541 & 0.975 & 0.879 & 0.987 & \textbf{0.941} \\
\cmidrule(r){1-9}
All category & 0.042 & 0.068 & 0.860 & 0.519 & 0.917 & 0.665 & 0.950 & \textbf{0.804} \\
\bottomrule
\end{tabular}
\end{table*}

We use F1-score~\cite{AccuracyFScoreROCFamilyDiscriminantMeasuresPerformanceEvaluation} and Matthews Correlation Coefficient~(MCC)~\cite{advantagesMatthewscorrelationcoefficientMCCF1scoreaccuracybinaryclassificationevaluation} for the evaluation. 
F1-score, as shown in Eq.~(\ref{eq:f1}) , is a common metric for binary classification.

\begin{equation}
    \text{F1-score} = \frac{2 \times \text{TP}}{2 \times \text{TP} + \text{FP} + \text{FN}}
  \label{eq:f1}
\end{equation}

As shown in Eq.~(\ref{eq:f1}), F1-score does not use prediction of true negative. 
Thus, when there is a large number of positive samples during inference, the performance can be significantly inflated by predicting all samples as positive.
For instance, in MVTec AD, with 1,258 positive samples and 467 negative samples in the test data, F1-score shows a high value of 0.844. This shows F1-score is suspicious when there is a bias in the test data.
Thus, we use not only F1-score but also MCC as evaluation metrics. 
MCC is reported to be adequate for binary classification, particularly for better consistency and less variance~\cite{MetricsMultiClassClassificationOverview}~\cite{GoodClassificationMeasuresHowFindThem}. 
MCC is shown in Eq.~(\ref{eq:mcc}).

\begin{equation}
    \text{MCC} = \frac{\text{TP} \times \text{TN} - \text{FP} \times \text{FN}}{\sqrt{(\text{TP} + \text{FP})(\text{TP} + \text{FN})(\text{TN} + \text{FP})(\text{TN} + \text{FN})}}
  \label{eq:mcc}
\end{equation}

MCC ranges from -1 to 1, where 1 indicates perfect prediction of all samples, -1 indicates incorrect prediction of all samples, and 0 indicates random prediction. 
In the previously mentioned example, MCC cannot be calculated because the denominator becomes zero.
Thus, in this study, we use both F1-score and MCC for evaluation.
The evaluation methods include assessing the performance for each product within each dataset, as well as the overall performance across the entire dataset.

\subsection{Evaluation of Results}
\subsubsection{Result of MVTec AD}
The results for MVTec AD are shown in Tab.~\ref{tab:result_mvtec}. 
The settings are as follows: ``Vanilla'' for ViP-LLaVA before fine-tune, ``w/o ICL'' for ViP-LLaVA after fine-tune without using an example during inference, ``ICL (RICES)'' for using a selected example image with the RICES algorithm during inference, and ``ICL (Ours)'' for using a selected example image with Eq.~(\ref{eq:ours}) during inference.
In each settings, results of F1-score and MCC are in a row.
From the table, we confirm that providing an example significantly improves performance.
This demonstrates the effectiveness of our framework. 
Additionally, compared to RICES, our selection algorithm achieves improvement in performance with an increase in MCC, demonstrating the effectiveness of our algorithm.

Next, for qualitative evaluation, the visualization of the model prediction is shown in Fig.~\ref{mvtecbb2}.
As shown in the figure, our approach can roughly detect defective locations, which means the model recognizes the defects in the image.
However, the model cannot detect multiple defects or logical defects, such as those in ``Cable''. This is due to the lack of variety in the training dataset.
Thus, further image collection and an enlarged training dataset are required for performance improvement.

Also, for some products like ``Hazelnut'', while our approach improved the performance, it is still insufficient for real-world conditions.
For ``Hazelnut'', the model detected thin parts as defective, indicating that the model does not fully leverage ICL.
Thus, providing detailed inspection criteria is necessary. 
It has been reported that increasing the number of examples improves ICL performance~\cite{ManyShotInContextLearning}~\cite{InContextLearningLongContextModelsInDepthExploration}.
Alternatively, further performance improvement is expected by proposing an optimal selection algorithm that selects multiple example images~(based on the query strategies, including those from Deep Active Learning~\cite{SurveyDeepActiveLearning}~\cite{BenchmarkingQueryStrategiesFutureDeepActiveLearning}).
\mvtecbb

Additionally, for all products, although coordinates are output, their positions deviate from the actual defective locations.
Indeed, pixel-level AUROC was 0.730, which is very low compared to the existing methods.
This is because the CrossEntropyLoss used for training uniformly calculates the loss for differences in token values.
For example, when the ground truth of the starting x-coordinate is 100, the loss is the same when the model outputs 101 and 900 (assuming the prediction probabilities are equal).
Thus, CrossEntropyLoss is not optimal for tasks requiring specific numerical outputs like coordinates. 
However, existing VLMs are trained with CrossEntropyLoss, meaning their outputs are text-based cannot be safely converted to floats~(with gradient flow intact), thus performance improvement is expected by constructing a multi-head VLM for defect detection and modifying the loss function to alternatives like Mean Squared Error or GIoU Loss.
\comparison

While ``Bottle'' has the same product in the training dataset, their performance is lower compared to ``Wood'', which also has the same product in the training dataset.
This is likely due to the significant differences in appearance between the images in the training data and those in MVTec AD, as shown in Fig.~\ref{comparison}.
However, despite the differences in appearance, ``Tile'' shows high performance, confirming the generalization capability for some products.
Also, to prevent the forgetting of knowledge acquired during pre-training when fine-tuning, it is necessary to use Parameter Efficient Fine-Tuning methods, such as Low-Rank Adaptation~\cite{LoRALowRankAdaptationLargeLanguageModels}, which forget less than fine-tuning~\cite{LoRALearnsLessForgetsLess}.

\begin{table*}[!tb]
\caption{Result of VisA.}
\label{tab:result_visa}
\centering
\begin{tabular}{lcccccccc}
\toprule
\multicolumn{1}{c}{\textbf{Settings}} & \multicolumn{2}{c}{\textbf{Vanilla}} & \multicolumn{2}{c}{\textbf{w/o ICL}} & \multicolumn{2}{c}{\textbf{ICL (RICES)}} & \multicolumn{2}{c}{\textbf{ICL (Ours)}} \\
\cmidrule(r){1-9}
\multicolumn{1}{c}{Product Name} & F1-score & MCC & F1-score & MCC & F1-score & MCC & F1-score & MCC \\
\midrule
Candle & N/A & N/A & 0.635 & \textbf{0.539} & 0.692 & 0.241 & 0.694 & 0.253  \\
Capsules & N/A & N/A & 0.599 & 0.415 & 0.841 & \textbf{0.513} & 0.809 & 0.389 \\
Cashew & N/A & N/A & 0.814 & 0.623 & 0.890 & 0.670 & 0.889 & \textbf{0.674} \\
Chewinggum & N/A & N/A & 0.921 & 0.758 & 0.921 & 0.758 & 0.935 & \textbf{0.804} \\
Fryum & N/A & N/A & 0.867 & 0.699 & 0.917 & \textbf{0.741} & 0.888 & 0.648 \\
Macaroni1 & N/A & N/A & 0.760 & \textbf{0.502} & 0.685 & 0.204 & 0.683 & 0.190 \\
Macaroni2 & N/A & N/A & 0.669 & \textbf{0.071} & 0.667 & N/A & 0.667 & N/A \\
PCB1 & N/A & N/A & 0.131 & 0.190 & 0.891 & \textbf{0.792} & 0.875 & 0.762 \\
PCB2 & N/A & N/A & 0.347 & 0.343 & 0.772 & \textbf{0.493} & 0.763 & 0.471 \\
PCB3 & N/A & N/A & 0.243 & 0.248 & 0.747 & 0.503 & 0.751 & \textbf{0.513} \\
PCB4 & N/A & N/A & 0.622 & 0.516 & 0.801 & 0.594 & 0.817 & \textbf{0.610} \\
Pipe Fryum & N/A & N/A & 0.870 & 0.726 & 0.920 & 0.744 & 0.929 & \textbf{0.774} \\
\cmidrule(r){1-9}
All category & N/A & N/A & 0.671 & 0.429 & 0.800 & \textbf{0.492} & 0.795 & 0.479 \\
\bottomrule
\end{tabular}
\end{table*}
\subsubsection{Result of VisA}

The results for VisA are shown in Tab.~\ref{tab:result_visa}. 
The table follows the same format as Tab.~\ref{tab:result_mvtec}.
From the table, it can be confirmed that the performance improves by using ICL in VisA as well, demonstrating the effectiveness of the proposed framework.
However, compared to RICES, our selection algorithm does not show significant improvement. 
This is because both RICES and our selection algorithm are based on similarity, which depends on the data distribution. 
Most of the products in VisA are too widely distributed~(e.g., ``Macaroni'', ``PCB''). 
Thus, proposing a more distribution-robust selection algorithm could potentially improve performance.
Also, it can be seen that the performance does not improve regardless of the presence of ICL when there are two or more products in the image, especially if those products are not aligned.
In fact, ``Macaroni1'', which is neatly aligned, shows higher qualitative and quantitative performance compared to ``Macaroni2'', which is randomly arranged.
This is likely due to the lack of training dataset that considers differences in product positions and orientations. 
Thus, performance improvement is expected by collecting fine-tuning data and performing data augmentation, such as rotation and flipping.
Simultaneously, it should be noted that for some products, positional shifts or orientation differences may be defined as defects.

For qualitative evaluation, the visualization of the model prediction is shown in Fig.~\ref{visa_bb}.
As shown in Fig.~\ref{visa_bb}, for products that have multiple objects like ``Candle'' or ``Capsules'', the model prediction gets worse.
As mentioned, our dataset is still insufficient for generalization because there are limited products and they are mostly single object. 
In addition, images with multiple objects are highly distributed compared to the images with single object, which influences the performance of ICL because the selection algorithms depend on the distribution.
\visabb

\section{Conclusion}
In this study, we propose a general visual inspection model based on a few images of non-defective or defective products along with explanatory texts serving as inspection criteria. 
For future work, further performance improvement is expected by collecting more images for fine-tuning. 
In this study, we enabled visual inspection using VLM by training on a dataset consisting of only 941 images, which is very small compared to the pre-training dataset of VLM. 
Another consideration is to construct the multi-head VLM and change of the loss function.
Furthermore, introducing the example image selection algorithm is another way for improvement. 
Specifically, existing algorithms are for selecting one example image for the inspection, so proposing an optimal selection algorithm for many example images improves model performance.
In addition, the proposed method is based on VLM, so by adding the rationale statements for the decision in the response, model explainability is expected to improve, and performance could be enhanced through multitasking.
Finally, in our study, we evaluate only on MVTec AD and VisA. However, more comprehensive benchmark such as MMAD~\cite{jiang2024mmad} is required.

\bibliographystyle{apalike}
{\small
\bibliography{main}}

\newpage
\section*{\MakeUppercase{Appendix}}

\section{Product Category}
As mentioned in Sec.~\ref{sec:dataset}, we fine-tuned LVLM using a diverse set of non-defective and defective images of various products collected from the web to enhance the visual inspection capabilities of LVLM. 
The product names in the dataset we used are shown in Fig.~\ref{category}. 
Note that we renamed the products during dataset construction after collection (e.g., CD → disk, carpet → textiles).
All the images used for training will be publicly available at https://github.com/ia-gu/Vision-Language-In-Context-Learning-Driven-Few-Shot-Visual-Inspection-Model.
\category

\section{Result of ViP-LLaVA Before Fine-tuning}
\resultwoft
An example of the prediction results on MVTec AD of ViP-LLaVA before fine-tuning is shown in Fig.~\ref{resultwoft} 
As illustrated, the vanilla ViP-LLaVA fails correct inspection, predicts both non-defective and defective products as defective. 
Moreover, the format of the response text is inconsistent, making it challenging to perform a consistent quantitative evaluation.
These results confirm the effectiveness of fine-tuning using our dataset. 
On the other hand, from the third result, it can be seen that the vanilla ViP-LLaVA is capable of describing the type and location of the defect with statements such as 'The defect is located at the center of the leather, and it appears to be a hole.' 
Therefore, by adding rationale statements to our dataset, it is suggested that the proposed method could output not only defective location coordinates but also the rationale statements for the judgment.

\section{Ablation Study of In-Context Learning for MVTec AD}
\begin{table*}[!tb]
\caption{Result of the comparison of In-Context Learning. All the results are by MCC. Each of the index means “1-pos” gives one defective example image, “1-neg” gives one non-defective example image, “2-pos-pos” gives two defective example images, “2-neg-neg” gives two non-defective example images, “2-pos-neg” gives one defective example image and one non-defective example image in a row, “2-neg-pos” is vice versa.}
\label{tab:mvtec_icl}
\centering
\begin{tabular}{lcccccc}
\toprule
\cmidrule(r){1-7}
\multicolumn{1}{c}{\textbf{Settings}} & \textbf{    1-pos    } & \textbf{    1-neg    } & \textbf{2-pos-pos} & \textbf{2-neg-neg} & \textbf{2-pos-neg} & \textbf{2-neg-pos} \\
\midrule
Bottle & 0.509 & 0.589 & N/A & N/A & 0.201 & 0.498 \\
Cable & 0.124 & 0.589 & 0.184 & 0.305 & 0.275 & 0.056 \\
Capsule & 0.185 & 0.353 & 0.006 & N/A & 0.305 & 0.164 \\
Carpet & 0.195 & 0.892 & 0.356 & 0.931 & 0.678 & 0.124 \\
Grid & 0.598 & 0.816 & 0.547 & 0.838 & 0.684 & 0.544 \\
Hazelnut & 0.428 & 0.198 & 0.519 & N/A & 0.105 & 0.495 \\
Leather & 0.578 & 0.939 & 0.343 & 1.000 & 0.734 & 0.603 \\
Metal Nut & 0.226 & 0.402 & 0.002 & 0.427 & 0.237 & 0.244 \\
Pill & 0.374 & 0.440 & 0.227 & 0.282 & 0.405 & 0.313 \\
Screw & 0.123 & 0.084 & 0.155 & 0.004 & 0.167 & 0.156 \\
Tile & 0.724 & 0.764 & 0.621 & 0.790 & 0.790 & 0.689 \\
Toothbrush & N/A & 0.484 & N/A & 0.496 & N/A & N/A \\
Transistor & 0.205 & 0.556 & 0.004 & 0.510 & 0.275 & 0.151 \\
Wood & 0.421 & 0.896 & 0.280 & 0.849 & 0.775 & 0.421 \\
Zipper & 0.260 & 0.482 & 0.280 & 0.418 & 0.410 & 0.242 \\
\cmidrule(r){1-7}
All category & 0.279 & 0.475 & 0.197 & 0.455 & 0.360 & 0.267 \\
\bottomrule
\end{tabular}
\end{table*}

In our main experiments, we confirmed that selectively providing a single example image during ICL in the evaluation improves performance.
Here, we use MVTec AD to verify the effectiveness of the proposed method by comparing results when example images are selected randomly. 
At the same time, we compare results when the number of example images is increased.

The experimental results are shown in Tab.~\ref{tab:mvtec_icl}.
From the table, it can be seen that the highest performance is achieved when a single non-defective example is provided randomly~(note that this is lower than the result of 'w/o ICL').
Additionally, simply increasing the number of provided examples does not improve performance; on the contrary, it decreases.
This indicates that in ICL, the influence of the examples is significant, and increasing the number of examples without considering their relevance to the query image leads to performance degradation.
Although our proposed method and RICES are algorithms specialized in selecting a single example image, the performance improvement expected from increasing the number of examples in ICL suggests that further improvements could be achieved by proposing an algorithm for selecting two or more examples.

\section{Visualization Results of All Products for MVTec AD and VisA}
The visualization of the model prediction for MVTec AD and VisA is shown in Fig.~\ref{mvtecbbfull} and Fig.~\ref{visabbfull}.
We can see the same tendency that when there are two or more products in the image, the performance decreases, and ICL does not work well.
The performance decrease is due to the lack of dataset diversity. Most images in our dataset contain a single product.
The reason why ICL does not work well is that when there are two or more products in each image of one category, the diversity within the category increases, and the algorithm for calculating similarity or distance fails to perform effectively.
Thus, for future work, it is noted that simply calculating similarity or distance may fail in specific domains.

\mvtecbbfull
\visabbfull

\end{document}